\documentclass[letterpaper]{article} 
\usepackage{aaai25}  
\usepackage{times}  
\usepackage{helvet}  
\usepackage{courier}  
\usepackage[hyphens]{url}  
\usepackage{graphicx} 
\usepackage{natbib}  
\usepackage{caption} 
\usepackage{algorithm}
\usepackage{algorithmic}

\usepackage{newfloat}
\usepackage{listings}
\DeclareCaptionStyle{ruled}{labelfont=normalfont,labelsep=colon,strut=off} 
\floatstyle{ruled}
\newfloat{listing}{tb}{lst}{}
\floatname{listing}{Listing}
\usepackage{multirow}
\usepackage{makecell}
\usepackage{url}
\usepackage{graphicx}
\usepackage{booktabs}
\usepackage{amsmath, bm}
\usepackage{amssymb}
\usepackage{mathtools}
\usepackage{amsthm}
\usepackage{listings}
\usepackage{xcolor}
\usepackage{adjustbox} 
\usepackage{subcaption}

\definecolor{customgreen}{HTML}{228B22}
\definecolor{customred}{HTML}{7d1211}

\title{A Deep Probabilistic Framework for Continuous Time Dynamic Graph Generation}
\author {
    Ryien Hosseini\textsuperscript{\rm 1},
    Filippo Simini\textsuperscript{\rm 2},
    Venkatram Vishwanath\textsuperscript{\rm 2},
    Henry Hoffmann\textsuperscript{\rm 1}
}
\affiliations {
    \textsuperscript{\rm 1}Department of Computer Science, University of Chicago\\

    \textsuperscript{\rm 2}Argonne National Laboratory\\
    ryien@uchicago.edu, fsimini@anl.gov, venkat@anl.gov, hankhoffmann@cs.uchicago.edu
}

\begin{document}

\maketitle

\begin{abstract}
Recent advancements in graph representation learning have shifted attention towards \textit{dynamic graphs}, which exhibit evolving topologies and features over time. The increased use of such graphs creates a paramount need for \textit{generative} models suitable for applications such as data augmentation, obfuscation, and anomaly detection. However, there are few generative techniques that handle continuously changing temporal graph data; existing work largely relies on augmenting static graphs with additional temporal information to model dynamic interactions between nodes. In this work, we propose a fundamentally different approach: We instead directly model interactions as a joint probability of an edge forming between two nodes at a given time. This allows us to autoregressively generate new synthetic dynamic graphs in a largely assumption free, scalable, and inductive manner. We formalize this approach as \textit{DG-Gen}, a generative framework for continuous time dynamic graphs, and demonstrate its effectiveness over five datasets.  Our experiments demonstrate that DG-Gen not only generates higher fidelity graphs compared to traditional methods but also significantly advances link prediction tasks.
\end{abstract}

\section{Introduction}
\label{sec:intro}

Recent research in graph representation learning focuses on  \textit{dynamic}\footnote{We use \textit{dynamic graph} and \textit{temporal graph} interchangeably.} graphs, or graphs which change over time~\citep{dynamic_graph_survey, kazemi_survey}. 
Dynamic graphs are necessary to model several interesting real world phenomena that cannot be captured by static graphs alone, as the underlying data distribution changes as a function of time. Examples include social networks, citation graphs, and financial transactions \citep{dynamic_graph_survey}.

Several methods learn in the discriminative regime \citep{ctdg_1, roland, ctdg_bastas2019evolve2vec, ctdg_ma2020streaming, ctdg_nguyen2018dynamic, jodie}. These models support a variety of downstream tasks such as link prediction, node and graph classification, etc. While considerable progress has been made in this discriminative paradigm, work in the \textit{generative} regime is nascent. Yet, generative models for temporal graphs are essential for tasks such as data augmentation, obfuscation of sensitive user data during dataset creation, and anomaly detection \cite{survey_graph_generation, generative_models_survey}. 

\paragraph{Problem Formulation.} 
The goal of continuous time dynamic graph (CTDG) generation is to learn a distribution over source graph(s) that are in a state of continuous evolution. This learning process must ensure that the statistical properties of graphs derived from this distribution closely resemble those of the source graphs. CTDGs are characterized by a series of timestamped \textit{events}, each involving a source and destination node and an edge feature vector. The range of possible events encompasses node additions and deletions, edge interactions, and feature evolution. 
Section \ref{sec:background_1} provides a formal mathematical definition of CTDGs. The challenge of useful CTDG generative modeling lies in learning a distribution to generate event sequences that both reflect and diverge from reference graphs --- achieving statistical similarity without exact duplication. This balance prevents information leakage and guarantees the uniqueness of the generated graphs, ensuring they are innovative reflections rather than replicas of the source data.

\paragraph{Limitations of Prior Approaches.}
We argue that current generative models for CTDGs are too reliant on existing static methods. Specifically, current approaches attempt to model temporal graphs either as a single static graph with temporal edges \cite{taggen, tigger} or as a set of static \textit{snapshots} representing the graph at discrete timestamps \cite{dymond}. Such methods then apply existing generative models for \textit{static graphs} (e.g. \citealp{gnn_gen_semantic_parsing_1, gnn_gen_semantic_parsing_2, gnn_gen_molecule_1, gnn_gen_molecule_2}) to generate sequences of evolving static graphs. 
However, this reliance on augmenting static graphs with discrete time representations falls short in addressing the intrinsic nature of temporal graphs: continuous and non-uniform evolution. This misalignment introduces the following challenges:
\begin{itemize}
    \item \textbf{Topological assumptions:} Prior work makes strong inductive assumptions about the source graph's underlying topology to fit it into a discrete series of static graphs. This leads to poor modeling of many real world datasets, where such assumptions may not hold.
    
    \item \textbf{Lack of inductive modeling:} Inductivity allows models to transfer knowledge about the underlying graph to unseen nodes \cite{graphsage}. Here, this refers to the ability of a model to create new synthetic nodes not seen during training. Most static representations of dynamic graphs rely on mapping node IDs and/or interaction timestamps directly to the generated graphs and are thus limited to purely \textit{transductive} modeling. Such an approach not only raises concerns such as node leakage from the source graph \cite{tigger}, but also inherently lacks the ability to model new nodes and thus restricts the model's utility in dynamic scenarios where the introduction of novel nodes is common.
    
    \item \textbf{Poor Scalability to Large Time Horizons:} Static representations of temporal graphs often necessitate the explicit computation of graph adjacency matrices. As these matrices grow in size, they can cause model divergence and substantially increase memory requirements. Consequently, most existing methods for graph generation are constrained to learning on small source graphs.
    \item \textbf{Inability to Model Network Features:} The approaches used in prior work are not able to incorporate existing edge features during training nor able to generate graphs with such synthetic features.
\end{itemize}

\paragraph{Our Key Insight.}
In contrast to prior work that fundamentally relies on augmenting static graph representations with temporal data, we propose a novel method that directly models temporal graphs. 
We directly model the \textit{interactions} (also referred to as \textit{temporal edges}) within a temporal graph, that is, a sequence of 4-tuples with each element in the tuple representing the source and destination nodes (along with any associated features), 
timestamp, and edge features, respectively. We define the occurrence of such a temporal edge as a joint probability of these elements. This probability is then factored into a product of conditional probabilities, each of which represent an appropriate statistical distribution whose parameters can be estimated using a \textit{deep probabilistic decoder} (See Figure \ref{fig:model}). This decoder can be paired with any encoder that maps temporal interactions to dense node embedding vectors. Once trained, this model directly generates synthetic interactions in an autoregressive manner.

\paragraph{Our Solution.}
We formalize the above idea into \textit{DG-Gen}\footnote{Source code available at \protect{\url{https://github.com/ryienh/DGGen}}} (\textbf{D}ynamic \textbf{G}raph \textbf{Gen}erative Network), 
an encoder-decoder framework for temporal graph generation. By directly modeling the probability of temporal interactions, DG-Gen addresses the challenges noted above: By learning parameters to invariant conditional probability distributions, we model topological evolution in a largely assumption-free manner. Additionally, our method models the probability of a single (or small mini-batched) temporal interaction at a time and consequently avoids encountering the data scalability (and memory) challenges associated with previous methods. Further, by learning to predict edge probabilities from a learned \textit{temporal} embedding, rather than a node ID, our model is fully inductive by default and thus generates graphs with unseen nodes and timestamps. Finally, our model generates output edge features of arbitrary length.

\paragraph{Contributions.}
We present DG-Gen, a generic \textit{generative framework} for CTDGs which models temporal graph interactions directly. Our framework combines an encoder module that creates a temporal latent space embedding from raw interaction data with a novel \textit{deep generative decoder}. This decoder learns a product of conditional probabilities which together form the joint probability of a graph interaction. Given the emerging state of research in this area, to our knowledge, only one existing model baseline, TIGGER-I \cite{tigger}, is capable of \textit{inductive} generation of CTDGs. We perform comprehensive experiments on a version of \textit{DG-Gen} which we show outperforms TIGGER-I on five diverse temporal graph datasets in generating high-fidelity and original synthetic data in an inductive manner. 

To further validate our approach, we adapt it for the \textit{discriminative} task of link prediction. This adaptation is valuable for two reasons: (1) it demonstrates the versatility of our autoregressive approach to model link prediction, and (2) it highlights the effectiveness of \textit{DG-Gen}'s learned embeddings, reinforcing the quality of our generative approach.

\section{Background and Related Work}
\label{sec:background}

\begin{table}[ht]
    \centering
    \small
    \begin{tabular}{l@{\hskip 5pt}c@{\hskip 5pt}c@{\hskip 5pt}c@{\hskip 5pt}c@{\hskip 5pt}c}
        Capability & \adjustbox{angle=55}{Dymond} & \adjustbox{angle=55}{TagGen} & \adjustbox{angle=55}{TIGGER} & \adjustbox{angle=55}{TIGGER-I} & \adjustbox{angle=55}{DG-Gen} \\
        \midrule
        Topology agnostic  & $\boldsymbol{\times}$ & $\boldsymbol{\checkmark}$ & $\boldsymbol{\checkmark}$ & $\boldsymbol{\checkmark}$ & $\boldsymbol{\checkmark}$ \\
        Inductive      & $\boldsymbol{\times}$ & $\boldsymbol{\times}$ & $\boldsymbol{\times}$ & $\boldsymbol{\checkmark}$ & $\boldsymbol{\checkmark}$ \\
        Time scalable  & $\boldsymbol{\times}$ & $\boldsymbol{\times}$ & $\boldsymbol{\checkmark}$ & $\boldsymbol{\times}$ & $\boldsymbol{\checkmark}$ \\
        Supports features & $\boldsymbol{\times}$ & $\boldsymbol{\times}$ & $\boldsymbol{\times}$ & $\boldsymbol{\times}$ & $\boldsymbol{\checkmark}$ \\
        \bottomrule
    \end{tabular}
    \caption{Comparison of DG-Gen and existing generative models temporal graphs for various capabilities.}
    \label{tab:comparison}
\end{table}

\subsection{Dynamic Graph Representations}
\label{sec:background_1}
Early work on dynamic graph representation learning focused on learning embeddings for \textit{discrete-time dynamic graphs} (DTDGs), or time sequences of static graphs. Formally, we can define a DTDG as a sequence $\mathcal{G} = \{G^{(1)}, G^{(2)}, ..., G^{(\tau)}\}$, of equally spaced static graph \textit{snapshots}, wherein each graph $G^{(t)}$ consists of adjacency matrix $A^{(t)}$, node feature matrix $N^{(t)}$, and edge feature matrix $E^{(t)}$. Such graph representations are useful for applications where data is captured at regularly spaced time intervals, such as sensors or monitoring systems \cite{kazemi2022dynamic}. 

\textit{Continuous-time dynamic graphs} (CTDGs) generalize DTDGs as timed event sequences encompassing node and edge additions, deletions, and node and edge feature evolution \citep{kazemi2022dynamic}. This paper concentrates on CTDGs, because of their applicability to real-world datasets (e.g., social networks, transportation systems).å
Unlike DTDGs, CTDGs offer a more practical representation for continuous-time data, as converting a CTDG to a DTDG can result in substantial information loss of events occurring at higher temporal resolutions than the snapshots \cite{kazemi2022dynamic}. 

Given a set of nodes $\mathcal{V} = \{1,...,n\}$, we represent a CTDG as a sequence of events over time, $\mathcal{G} = \{x(t_1), x(t_2),..., x(t_{\tau})\}$ where each event $x(t_i)$ occurring at timestamp $t_i$ with $i \in \{1,..., \tau\}$ represents a node or edge event (that is, creation, deletion, feature transformation). Specifically, each event is characterized by the following attributes: $x(t_i) = (src, dst , t_i, \mathbf{e}_{src, dst}(t_i))$, where $src, dst \in \mathcal{V}$ are the source and destination nodes of the interaction and are represented by a unique scalar valued \textit{node ID} $\in \{1, ..., n\}$, and $\mathbf{e}_{src, dst}(t_i)$ 
is a directed edge between source node $src$ and destination node $dst$ at time $t_i$ and is represented by its feature vector. 

\subsection{Discriminative Models for CTDGs}
\label{sec:background_2}
Representation learning for CTDGs is a rapidly growing area. Early methods, as in \citealp{ctdg_bastas2019evolve2vec} and \citealp{nguyen2018continuous}, integrate the temporal aspects into a static graph representation as \textit{temporal edges}. Then, random walk methods (e.g. \citealp{node2vec, deepwalk}) learn representations. Other approaches \citep{jodie, trivedi2017know, ctdg_ma2020streaming} use recurrent neural networks (RNNs) \citep{rnn} to incorporate temporal information. Additional work, such as \citealp{tgor} and \citealp{dyngraph}, combines random walks with auto-encoders \cite{auto_enc} and generative adversarial networks \cite{gan}, to create embeddings that capture long-term temporal dependencies. 

Perhaps one of the most significant advances in representation learning for CTDGs is TGN \citep{tgn}, which proposes a generic framework for inductive learning dynamic graphs. TGN uses a memory module of seen nodes and edges to iteratively update unseen nodes and interactions (edges). \citealp{tgn} shows that many other CTDG representation learning approaches such as DyRep \citep{ctdg_dyrep}, Jodie \citep{jodie}, and TGAT \citep{ctdg_1} 
are specific instances of this framework. However, other recent works --- e.g., CAWN \citep{cawn} ---
demonstrate comparable results applying static methods to CTDGs. For completeness, we compare our method's performance in the discriminative regime (link prediction), to such methods as well.

\subsection{Generative Models for Dynamic Graph Data}
\label{sec:background_3}

Generative models for dynamic graphs learn the evolving probability distribution of graph data over time. Given an input graph $\mathcal{G}^{\text{source}}$, the goal is to produce a model that learns $p(\mathcal{G}^{\text{source}})$ such that sampling from this distribution generates a novel synthetic graph $\mathcal{G}^{\text{synth}}$ that is statistically similar to $\mathcal{G}^{\text{source}}$. Evaluation of these models typically involves comparing the synthetic graphs to the source graphs in terms of classical graph properties (e.g., mean degree, betweenness centrality \cite{graph_metrics}), assessed across various snapshots $\mathcal{G} = \{\mathcal{G}_{t_i},...\mathcal{G}_{t_n}\}$, regardless of DTDG/CTDG distinction. Originality is measured by the degree of edge overlap between $\mathcal{G}^{\text{source}}$ and $\mathcal{G}^{\text{synth}}$. Our work adheres to these evaluation standards, as detailed in Section \ref{sec:experiments}.

\paragraph{Generating DTDGs.} Early work in temporal graph generation focuses on DTDGs,  learning spatio-temporal embeddings of the graph snapshots and then adapting generative methods for static graphs. These include approaches based on \textit{random-walks} \cite{netgan, stggan}, \textit{variational autoencoders (VAEs)} \cite{vae_gen_1, d2g2}, and \textit{generative adverserial networks (GANs)} \cite{gcn_gan, graph_gan, grl, tggan}. While such methods show promise in generating graphs in the DTDG regime, they are unable to model continuous graph evolution due to the temporal resolution loss inherent to representing continuous graphs as snapshots \cite{tgn, survey_graph_generation}.

\paragraph{Generating CTDGs.} Investigations into generative models that learn from continously evolving graphs --- represented as CTDGs --- remain notably scarce, but we identify three distinct methodologies that can be applied to CTDG datasets. Each, however, significantly relies on representations akin to either DTDGs or static graphs which introduces substantial limitations as described below. Table \ref{tab:comparison} summarizes each of these approaches' ability to address the four challenges discussed in Section \ref{sec:intro}.

TagGen \citep{taggen} models a CTDG by transforming it into an equivalent static representation: 
a single static graph with temporal edges. Such an approach presents multiple challenges. First, temporal timestamps are modeled as discrete random variables, inhibiting the ability to generate interactions with unseen timestamps. Similarly, graph generation is reliant on a mapping from source graph node IDs, which renders both inductive modeling and feature modeling impossible. This leads to a critical issue where the generated graphs do not just statistically resemble the source graph; they predominantly consist of direct replications of the original data. As reported by \citealp{tigger}, up to 80\% of the generated graph's edges overlap with the source graph. This suggests that observed statistical resemblance is a consequence of duplicating the original dataset, undermining the goal of generating new graphs. Finally, this single static representation causes memory and performance issues for graphs with a larger temporal horizon, as an adjacency matrix of this static graph must be explicitly represented and inverted: In practice, this method is limited to graphs of the order of 200 timestamps \cite{tigger}. Conversely, real-world datasets such as the one used in our experiments are on the order of hundreds of thousands to millions of timestamps (See Appendix A for more details).

Dymond~\cite{dymond} introduces a generative model based on 3-node motifs, employing an approach that treats motif types as having a constant arrival rate. The model learns parameters to accurately represent this rate for each motif type. While new motifs are sampled at discrete intervals, akin to DTDG generation, their associated timestamps are derived from a learned continuous inter-event distribution. This methodology effectively produces a CTDG, but it presupposes that motifs appear at constant rates and, once formed, remain static throughout time. Consequently, unlike TagGen, empirical results are distinct but often markedly deviate from the statistical properties of the source graph. Moreover, the reliance on generating graphs over discrete intervals inherently restricts the method to only mapping node IDs observed during training, thereby preventing feature modeling or inductive graph generation. This approach also explicitly models graph snapshot adjacency matrices. Thus, \citealp{tigger} finds that this approach is likewise limited to graphs of the order of 200 timestamps.  

Recently, \citealp{tigger} introduced TIGGER, a CTDG generative method that addresses prior issues by decomposing graphs into temporal random walks, modeled through a recurrent neural network. These walks are then autoregressively sampled and merged to create synthetic graphs. Compared to Dymond, it makes fewer topological assumptions and scales better than other transductive methods.

However, such an approach faces challenges, particularly in its inductive variant, TIGGER-I. Node embeddings are derived by first converting the input graph into a static form---akin to TagGen---so that a static method (in this case GraphSAGE~\citep{graphsage}) can be applied. Then a static graph generator \cite{wgan} is trained. This dependence on static graph methodologies reintroduces the scalability issues discussed above and thus limits TIGGER-I's ability to model large time horizons. The authors acknowledge the issue, concluding that due to reliance on these static methods, the inductive version of their model is ``challenging to train on large graphs" \cite{tigger} and thus limit their inductive evaluation to graphs on the order of hundreds of timestamps. Additionally, both TIGGER and TIGGER-I model walks as node-time sequences, rendering them incapable of incorporating node or edge features during training and generation.

\paragraph{Inductive versus Transductive Graph Generation.}
Inductive graph generation differs from transductive graph generation in both its approach and applicability. Transductive models learn from a fixed set of nodes and edges, limiting their ability to generalize to new, unseen elements during inference. This restricts their use in dynamic environments where the underlying source graph continuously evolves.

In contrast, inductive graph generation allows the model to generate new nodes and edges by learning patterns in the underlying graph topology. This is advantageous in real-world applications where graphs frequently change. Inductive models are more flexible and generalizable, making them better suited for adapting to new data \cite{graphsage}.

Our work focuses on the inductive setting for continuous time dynamic graph generation. To the best of our knowledge, TIGGER-I is the sole generative CTDG method supporting inductive modeling and thus serves as our evaluation baseline in Section \ref{sec:experiments}. It is important to note that direct comparisons between inductive and transductive settings may be misleading, as the two approaches operate under fundamentally different assumptions and objectives.

\paragraph{Advancing CTDG Generation.} In the following section, we propose a novel generative CTDG method based on direct event probability modeling. Contrary to previous approaches, our method eschews reliance on static graphs or snapshots. This results in a model adept at inductively and scalably generating synthetic graphs, while also handling node and edge features of any size.


\begin{figure}[t]
\centering
\includegraphics[width=0.46\textwidth]{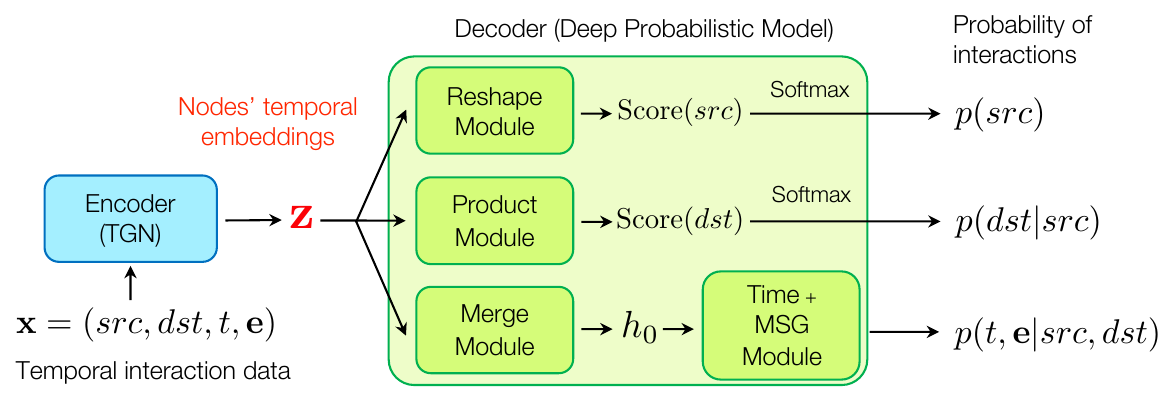}
\caption{Overview of DG-Gen's architecture and internal modules, described in Section~\ref{sec:model}.
}
\label{fig:model}
\end{figure}

\section{The Dynamic Graph Generative Network}
\label{sec:model}

DG-Gen (\textbf{D}ynamic \textbf{G}raph \textbf{Gen}erative Network) 
is a generative framework for Continuous-Time Dynamic Graphs (CTDGs) that computes temporal edge probabilities, enabling autoregressive edge sampling. DG-Gen is based on an encoder-decoder architecture (see Figure \ref{fig:model}). 

The encoder is a flexible component of our framework, capable of utilizing any existing method that maps raw interaction data to node embeddings. In our experiments, we employ the Temporal Graph Network (TGN) \cite{tgn} as the encoder due to its proven efficacy in capturing temporal dynamics. The decoder---a deep probabilistic model for tabular data, and a novel contribution of this paper---uses said embeddings to model the interactions.

\subsection{Encoder}
The Encoder processes raw temporal interaction data to generate temporal embeddings for each node in the input graph. A temporal embedding is a dense vector representation of a node's state at a specific point in time, informed by the node's current and past interactions, as well as the states of its temporal neighbors.
We specifically use TGN \cite{tgn}, which is composed of two main modules: the \textit{memory module} and the \textit{embedding module}. The memory module updates each node's \textit{memory} --- a dense vector storing relevant information about the node's past interactions. It aggregates the node's interactions from the previous temporal batch, combines them with the current memory, and produces an updated memory.
The embedding module, a transformer-based \citep{shi2020masked} convolution model, integrates the memories of a node’s temporal neighborhood along with edge features and temporal encodings to generate the final temporal embedding.

\begin{figure*}[t]
\centering
\includegraphics[width=0.76\textwidth]{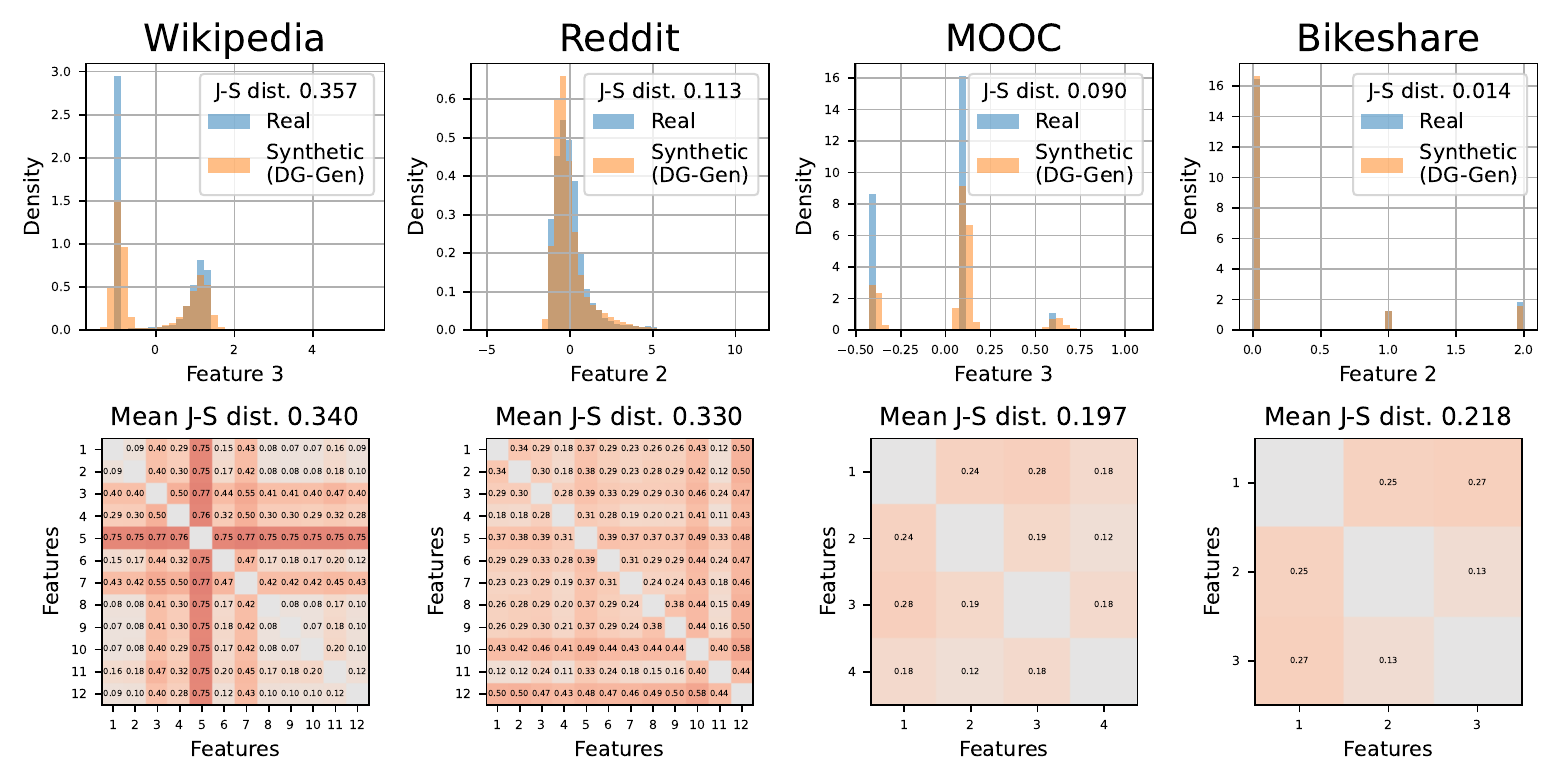}
\caption{Top panels: Histograms of one randomly selected edge feature distribution per dataset of the real data (blue) and of the synthetic data generated by DG-Gen (orange). Bottom panels: Jensen-Shannon distances between the 2-dimensional histograms (real and synthetic) of the joint distributions of feature pairs.}
\label{fig:gen_results}
\end{figure*}

\subsection{Decoder}
\textbf{Interaction Probability Factorization.}
The DG-Gen Decoder (Figure \ref{fig:model}) uses the temporal embeddings produced by the Encoder to create a probabilistic model of the raw interactions. 
As described in Section \ref{sec:background_1}, a raw interaction (or temporal edge) of a CTDG is specified by the following attributes: the source node ID ($src$), the destination node ID ($dst$), the time of the interaction ($t$), and the edge feature vector ($\mathbf{e}$) of size $n$. 
A key aspect of DG-Gen is the factorization of the joint probability of a raw interaction:
$p(src, dst, t, \mathbf{e}) = p(src) p(dst|src) p(t, \mathbf{e}|dst, src)$,
where this joint probability is expressed as the product of three distinct terms:
the probability of a node to be a source, $p(src)$, the conditional probability of a node to be a destination of an interaction with a given source node, $p(dst|src)$, and the conditional probability $p(t, \mathbf{e}|dst, src)$ that an interaction between a given source and destination node happens at time $t$ and has edge features $\mathbf{e} = \{e_1, ..., e_n\}$. 
The probability of the time and edge features given the source and destination nodes can be further decomposed as 
$p(t, \mathbf{e}|dst, src) = p(t |dst, src) p(e_1 |t, dst, src) ...  p(e_n |e_{n-1}, ..., e_1, t, dst, src)$. 

This factorization allows for a structured, modular approach to modeling complex temporal interactions, distinguishing DG-Gen from previous methods that rely on augmenting static graphs with temporal data.

\textbf{Decoder Overview. } The above probabilities are defined using appropriate statistical distributions whose parameters are estimated using neural networks:  
$p(src)$ is a Categorical distribution over the possible source nodes, where the probability of each node is estimated from the temporal embeddings using the Reshape Module (defined below) followed by a Softmax function. 
Similarly, $p(dst|src)$ is a Categorical distribution over the possible destination nodes given a source node, where the probability of each node is estimated from the temporal embeddings combined with the embedding of the (known) source node using the Product Module (defined below) followed by a Softmax function. 
$p(t |dst, src)$ is 
an Exponential distribution for the {\em inter-event time}, that is, the time difference between the current interaction (between the specified source and destination nodes) and the previous interaction. The actual (absolute) time is then obtained by adding the inter-event time to the absolute time of the previous interaction, i.e. performing a cumulative sum of the chronologically ordered inter-event times. 
The parameter of the Exponential distribution is obtained as the first output of the Time+MSG Module (defined below), a sequence model that is initialized combining the source's and destination's node embeddings via the Merge Module (defined below). $p(e_i |t, dst, src)$ is a Categorical or Gaussian Mixture Model (GMM) distribution for the $i$-th edge feature, depending on whether $e_i$ is a categorical or numerical variable, respectively. The distribution parameters (logits or means, standard deviations and mixture weights) are obtained as the $i$-th output of the Time+MSG Module. 
The following describe the above-mentioned Modules.

\textbf{Reshape Module.} This module takes a node embedding vector $\mathbf{z}_{src}$ as input and returns a scalar representing the node's score. In our experiments, it is defined as a linear transformation followed by a ReLU nonlinearity and another linear transformation to produce a one-dimensional output:
$\text{Reshape}(\mathbf{z}) = \mathbf{W}_f \cdot ReLU(\mathbf{W}_{src} \cdot \mathbf{z})$

\textbf{Product Module.} This module takes two node embedding vectors as input---$\mathbf{z}_{src}$  and $\mathbf{z}_{dst}$---and returns a scalar representing the interaction score between them. In our experiments the score is obtained performing a linear transformation of the two input embeddings, adding the resulting vectors, applying an element-wise ReLU nonlinearity, and performing a final linear transformation to a one-dimensional output: 
$\text{Product}(\mathbf{z}_{src}, \mathbf{z}_{dst}) = \mathbf{W}_f \cdot ReLU(\mathbf{W}_{src} \cdot \mathbf{z}_{src} + \mathbf{W}_{dst} \cdot \mathbf{z}_{dst})$

\textbf{Merge Module.} This module takes two embedding vectors, $\mathbf{z}_{src}$  and $\mathbf{z}_{dst}$, as input and returns a combined vector, which is used as input to the Time+MSG Module. Our experiments obtained it by performing a linear transformation of the two input embeddings, applying an element-wise ReLU nonlinearity, adding the resulting vectors and applying another nonlinearity followed by 
a final linear transformation to obtain the vector $\mathbf{h}_0$: 
$\text{Merge}(\mathbf{z}_{src}, \mathbf{z}_{dst}) = \mathbf{W}_f \cdot ReLU(ReLU(\mathbf{W}_{src} \cdot \mathbf{z}_{src}) + ReLU(\mathbf{W}_{dst} \cdot \mathbf{z}_{dst}))$

\textbf{Time+MSG Module.} This sequence model takes the output of the Merge Module, $\mathbf{h}_0$, and produces $n + 1$ outputs, where the $i$-th output corresponds to the parameters of the distribution describing the $i$-th edge feature $e_i$ (for $i > 0$) or the time $t$ (for $i = 0$). 
For $i = 0$ the output is passed through a linear layer to obtain 
a one-dimensional scalar which is converted to a positive value using a Softplus function, representing the inter-event time. 
For $i > 0$, the output is converted to a one-dimensional score if the edge feature $e_i$ is a categorical variable described by a Categorical (Multinomial) distribution. Otherwise $e_i$ is a numerical variable, so it is converted to $3 \cdot m$ numbers corresponding to the mean, standard deviation and weight of the $m$ components of a Gaussian Mixture Model.
Our experiments use a GRU \citep{gru} recurrent neural network and the Merge Module's output $\mathbf{h}_0$ is the initial hidden state of the network. 

\subsection{Training}
DG-Gen is trained using the Adam optimizer \citep{adam} to minimize the observed interaction data's negative log-likelihood, which can be computed as the sum of the log-likelihoods of the conditional probabilities parameterized by the Decoder modules: 
$loss = -\sum_{i} [\log p(src_i) +  \log p(dst_i|src_i) + \log p(t_i, \mathbf{e}_i|dst_i, src_i) ]$ where $src_i$, $dst_i$, $t_i$, $\mathbf{e}_i$ are source, destination, time and features of interaction $\mathbf{x}_i$. 
To accelerate training for large graphs, a random subset of nodes (usually of the order of two times the batch size or larger) is sampled when computing the scores of sources and destinations. 
Additionally, to improve numerical stability during training of the Time+MSG Module, a Gaussian random noise with zero mean and small standard deviation ($\sim 0.1$) is added to the numerical variables to mitigate the potential instability caused by discontinuous distributions, e.g., for the Wikipedia dataset's edge features (see Figure~\ref{fig:gen_results}). The standard deviation of this stabilizing noise is iteratively reduced to small values ($\sim0.001$) during training to gradually recover the original shapes of the feature distributions. 

\subsection{Inference}
\label{sec:method_link_pred}

\textbf{Generation. }
DG-Gen generates synthetic dynamic graphs by sampling interactions and their attributes from the distribution $p(src, dst, t, \mathbf{e}) = p(src) p(dst|src) p(t, \mathbf{e}|dst, src)$. First, a batch of source nodes is sampled from $p(src)$ based on current embeddings; second, a destination is sampled for each source using $p(dst|src)$; third, the embeddings of each source-destination pair are combined to instantiate the parameters and sample from the distributions of the $(t, \mathbf{e})$ variables. Initially, DG-Gen starts with an empty graph and all nodes have empty memory. After generating a batch of synthetic interactions, memories and neighbor lists are updated and new node embeddings are computed.

\textbf{Link Prediction. }
 DG-Gen learns the factored joint probability $p(src, dst, t, \mathbf{e}) = p(src) p(dst|src) p(t, \mathbf{e}|dst, src)$. The conditional probability $p(dst|src)$ directly corresponds to the established discriminative task of \textit{link prediction}, where the goal is to predict the likelihood of an edge (link) forming between a given source node and a destination node. Thus, DG-Gen can perform link prediction without further training or fine-tuning. In addition to generative analysis, our experiments in Section \ref{sec:experiments} compare the performance of DG-Gen to several state-of-the-art baselines for link prediction, further validating the quality of our approach.


\begin{table}[ht!]
\small
\centering
\setlength{\tabcolsep}{1mm} 

\begin{tabular}{lp{2.5cm}rrrrrr}
\toprule

Dataset & Model & CC & MD & NC & PLE & WC \\

\hline
\multirow[c]{2}{*}{Wikipedia} 
 & DG-Gen (no mem.)     & 0.723           & 0.013                  & \bfseries 1  & 0.191              & 1        \\
 & DG-Gen (no noise)    & 0.701           & 0.005                  & 2            & 0.185                & \bfseries 0        \\
 & TIGGER-I             & 0.992           & 0.007                  & 2            & 0.311                 & \bfseries 0        \\
 & DG-Gen               & \bfseries 0.656 & \bfseries 0.004        & \bfseries 1  &  \bfseries 0.099      & \bfseries 0        \\
\hline
\multirow[c]{2}{*}{Bikeshare} 
 & DG-Gen (no mem.)     & 0.305            & 0.170        & 14             & 9.983        & 4        \\
 & DG-Gen (no noise)    & \bfseries 0.297  & 0.200          & 15            & 10.206       & \bfseries 3        \\
 & TIGGER-I             & 0.857            & 0.280          & 21            & 10.378       & 15        \\
 & DG-Gen               & \bfseries 0.298  & \bfseries 0.125 & \bfseries 12  & \bfseries 9.929 & \bfseries 3        \\

\hline
\multirow[c]{2}{*}{MOOC} 
 & DG-Gen (no mem.)     & 0.486            & 0.611            & 2            & 5.328        & \bfseries 3        \\
 & DG-Gen (no noise)    & 0.449            & 0.425            & \bfseries 1  & 4.714        & \bfseries 3        \\
 & TIGGER-I             & 0.500            & 0.444            & 2            & 6.692       & 5        \\
 & DG-Gen               & \bfseries 0.430  & \bfseries 0.416  & \bfseries 1  & \bfseries 4.607  & \bfseries 3        \\

\hline
\multirow[c]{2}{*}{Reddit} 
 & DG-Gen (no mem.)     & 0.133             & 0.043            & 5       & 9.976       & \bfseries 1        \\
 & DG-Gen (no noise)    & 0.089             & 0.037            & 5       & 8.248       & 2        \\
 & TIGGER-I             & 0.242             & 0.059            & 5       & 10.120      & \bfseries 1        \\
 & DG-Gen               & \bfseries 0.066   & \bfseries 0.033  & \bfseries 4  & \bfseries 5.771       & \bfseries 1       \\

\hline
\multirow[c]{2}{*}{LastFM} 
 & DG-Gen (no mem.)     & 0.760           & 0.034             & \bfseries 1  & 2.076        & 2        \\
 & TIGGER-I             & 0.849           & 0.063             & 4       & 6.290        & 18        \\
 & DG-Gen               & \bfseries 0.681 &  \bfseries 0.027   & \bfseries 1  & \bfseries 1.282        & \bfseries 1        \\
\bottomrule

\end{tabular}

\caption{Median absolute error of metrics between real and synthetic data, quantifying the distance between real and generated properties. Best model is \textbf{bolded} (lower is better).}
\label{tab:results_table}
\end{table}

\section{Experiments}
\label{sec:experiments}
\textbf{Datasets and Experimental Setup.}
We evaluate our model using five CTDG datasets: Reddit, Wikipedia, MOOC, LastFM \citep{jodie} and Capital Bikeshare. The first four are established datasets in dynamic graph learning while Bikeshare is a novel dataset (see Appendix A for more information). 
Our experimental setup is split into two major parts. We first evaluate DG-Gen on the  inductive temporal graph generation task. The goal is to produce synthetic graphs that statistically align with real dynamic graphs seen during training. We assess these synthetic graphs using various statistical metrics, focusing on edge features and graph topology. Subsequently, we demonstrate how DG-Gen effectively learns representations for discriminative tasks by directly applying its learned probabilities to link prediction. Further details about reproducibility and experimental details are in Appendices B and C, respectively.

\textbf{Evaluation Metrics.}
To evaluate synthetic graph quality, we adopt the standard approaches discussed in Section \ref{sec:background_3} that ensure statistical similarity and originality of the generated graph. 
First, DG-Gen produces features that preserve the statistical properties of the source graph. We evaluate this ability by calculating distribution histograms for all interaction features and joint distribution histograms for all pairs of features. We report the mean of the Jensen-Shannon distance of these histograms as it is a standard measure of similarity between probability distributions \cite{lin1991divergence}. 
Second, synthetic graphs must retain the topological characteristics of the source graph. Following the generative baseline established by TIGGER-I, we partition each CTDG into discrete-time snapshots and assess the median absolute error across various graph statistics: closeness centrality (CC), mean degree (MD), number of components (NC), power law exponent of degree distribution (PLE), and wedge count (WC). Discussion of CTDG discretization can be found Appendix D.
Finally, synthetic graphs should be \textit{original}, that is, not simply copy the original source graph. \citealp{tigger} quantifies this idea by evaluating the fraction of overlapping edges between the source and synthetic graphs, that is $\frac{|\mathcal{E}^{source} \cap \mathcal{E}^{synthetic}|}{|\mathcal{E}^{source}|}$. We empirically find both DG-Gen and TIGGER-I exhibit no edge overlap; thus, we exclude this metric from our results.

For the link prediction task, we report Average Precision (AP) as our main evaluation metric. Following \citealp{better_metrics}, we use \textit{inductive sampling} for a more robust assessment of model performance on unseen edges. Additional results, including AUROC for inductive sampling and AP and AUROC for standard sampling, are in Appendix E.

\textbf{Baselines.}
To the best of our knowledge, TIGGER-I is the only method for inductive CTDG generation, so we compare DG-Gen's generated graph topology to TIGGER-I (see Appendix F for TIGGER-I training details). 
However, as TIGGER-I cannot generate synthetic edge features, we compare DG-Gen's synthetic edge features to those of the original source graph. We hope these results serve as a baseline for future methods generating CTDG edge features. For the inductive link prediction task, we benchmark against several models discussed in Section \ref{sec:background_2}: Jodie, DyRep, TGAT, TGN, and CAWN. Existing results from \citealp{better_metrics} are used for all but the novel Bikeshare dataset, where we train TGN and CAWN (see Appendix E).

\begin{table}[t]
\small
\centering
\begin{tabular}{lll}
    \toprule
    J-S dist. & \bfseries Single feature & \bfseries Feature pair \\
    \midrule
    \bfseries Wikipedia & 0.208 $\pm$ 0.154 & 0.340 $\pm$ 0.230 \\
    \bfseries Bikeshare & 0.107 $\pm$ 0.103 & 0.218 $\pm$ 0.059 \\
    \bfseries MOOC & 0.108 $\pm$ 0.087 & 0.197 $\pm$ 0.052 \\
    \bfseries Reddit & 0.191 $\pm$ 0.093 & 0.330 $\pm$ 0.111 \\
    \bottomrule
\end{tabular}
\caption{Mean and standard deviation of Jensen-Shannon distances between real and DG-Gen synthetic feature distributions, covering both single features and feature pairs.}
\label{tab:feats_table}
\end{table}

\subsection{Results} 
\textbf{Graph Generation.} A trained DG-Gen model can generate a synthetic temporal graph with statistical properties similar the real temporal graph seen during training. 
To demonstrate, we generate a synthetic graph with the same number of interactions as the test partition for each of the five datasets. We first evaluate DG-Gen's ability to produce realistic edge features over time. Figure~\ref{fig:gen_results} shows an example feature histogram per dataset, excluding LastFM, which lacks edge features.
Additionally, the figure shows Jensen-Shannon distances between real and synthetic 2D histograms of joint feature distributions. The mean values of all distances
are summarised in Table \ref{tab:feats_table}. These results show that DG-Gen (1) produces edge features with similar distributional properties as the underlying source graph for both categorical and numerical variables --- even with multimodal distributions --- and (2) captures the relationship between feature pairs. To the best of our knowledge, DG-Gen is the first model capable of generating edge features in CTDGs, preventing direct comparison with existing baselines. 

Regarding the analysis of graph topological properties, Table~\ref{tab:results_table} shows that DG-Gen outperforms TIGGER-I in most metrics across all datasets, notably excelling in closeness-centrality~\cite{closeness}.  
This metric's performance, particularly underscored in the LastFM dataset -- the largest among those evaluated, as detailed in Appendix A
--- attests to DG-Gen's adeptness at capturing extensive range dependencies within graph structures even at large scales.

\begin{figure}[t]
    \centering
    \includegraphics[width=1\linewidth]{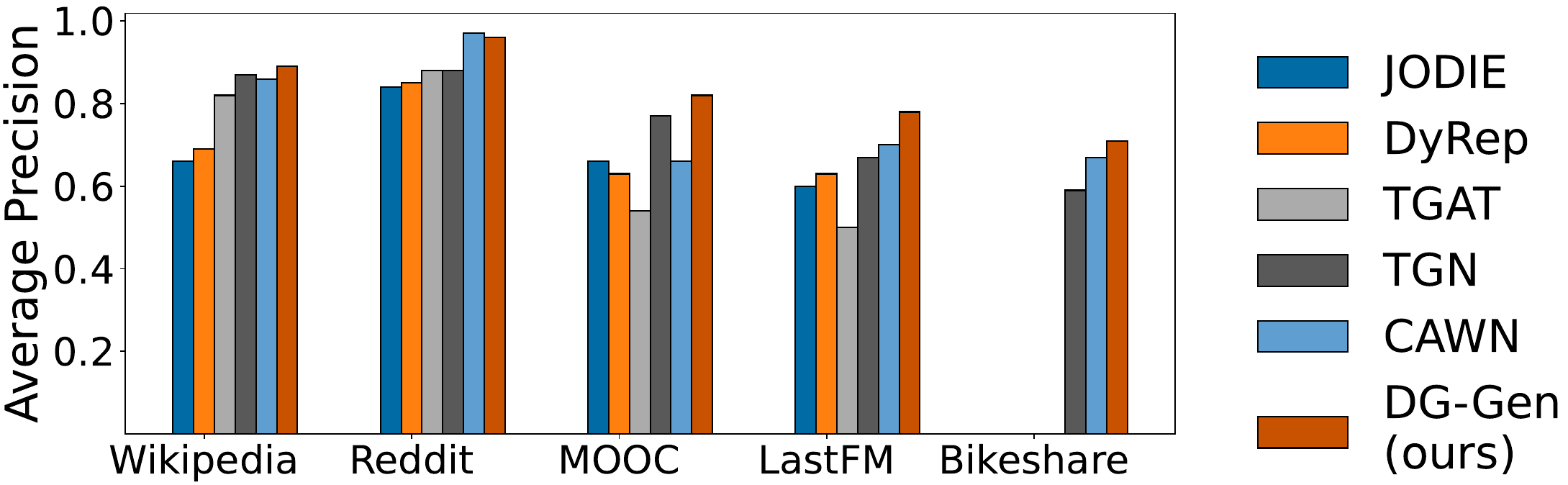} 
    \caption{Average Precision for DG-Gen and baselines on link prediction via inductive sampling.}
    \label{fig:lp_results}
\end{figure}

\textbf{Inductive Link Prediction.} 
Figure \ref{fig:lp_results} reports the AP score using inductive sampling. Appendix E includes AUC performance with inductive sampling and AP and AUC scores with standard sampling. 
DG-Gen outperforms the baselines on all but the Reddit dataset, where it matches CAWN. Notably, DG-Gen outperforms TGN while using the latter as an encoder, implying that the deep probabilistic decoder leads to better representation learning than TGN alone.


\textbf{Ablations.}
\label{sec:ablation}
We compare DG-Gen, which utilizes graph attention layers, with an otherwise identical model lacking these layers. The results, labeled \textit{DG-Gen (no mem.)} in Table \ref{tab:results_table}, show that omitting attention reduces performance across all datasets and metrics, with the most significant drop in closeness centrality (CC), a measure of long-range topological fidelity.
This finding underscores the importance of the attention module in capturing long-term global information. 

As detailed in Section \ref{sec:model}, we introduce random noise during training to stabilize the Time+MSG module's convergence while learning the edge feature distribution. Table \ref{tab:results_table} compares this approach with a noise-free training scheme, labeled \textit{DG-Gen (no noise)}. The results show a decline in data quality without noise, particularly for discrete or mixed feature distributions.

\section{Conclusion}

We have introduced \textit{DG-Gen}, the first inductive generative framework for CTDGs that handles arbitrarily-sized edge features. Unlike prior methods, our approach directly models the probability of edge events, eschewing reliance on static graph snapshots. 
DG-Gen generates graphs that, while original, are statistically similar to user-provided graphs. This underlines the framework's ability to replicate essential statistical properties, ensuring synthetic graphs are realistic proxies for original data.
Additionally, our method learns representations of dynamic graphs that outperform existing state-of-the-art 
approaches for tasks like link prediction. 

\section*{Acknowledgments}
This research used resources of the Argonne Leadership Computing Facility, a U.S. Department of Energy (DOE) Office of Science user facility at Argonne National Laboratory and is based on research supported by the U.S. DOE Office of Science-Advanced Scientific Computing Research Program, under Contract No. DE-AC02-06CH11357. Additional funding support comes from the National Science Foundation (CCF-2119184 CNS-2313190 CCF-1822949 CNS-1956180).

\bibliography{aaai25}

\end{document}